\documentclass[conference]{IEEEtran}
\IEEEoverridecommandlockouts

\usepackage{cite}
\usepackage{amsmath,amssymb,amsfonts}
\usepackage{algorithm}
\usepackage{algorithmic}
\usepackage{graphicx}
\usepackage{textcomp}
\usepackage{xcolor}

\def\BibTeX{{\rm B\kern-.05em{\sc i\kern-.025em b}\kern-.08em
    T\kern-.1667em\lower.7ex\hbox{E}\kern-.125emX}}

\newcommand{\armse}[2]{#1 \: {\scriptscriptstyle \pm #2}}

\begin{document}
\title{Robust Unscented Kalman Filtering via Recurrent Meta-Adaptation of Sigma-Point Weights}

\author{\IEEEauthorblockN{Kenan Majewski*, Michał Modzelewski, Marcin Żugaj, Piotr Lichota }
\IEEEauthorblockA{\textit{Institute of Aeronautics and Applied Mechanics} \\
\textit{Warsaw University of Technology}\\
Warsaw, Poland \\
*kenan.majewski.dokt@pw.edu.pl}
\thanks{This work has been submitted to the IEEE for possible publication.
Copyright may be transferred without notice, after which this version may
no longer be accessible.
}
\vspace{-10pt}
}

\maketitle

\begin{abstract}
  The Unscented Kalman Filter (UKF) is a ubiquitous tool for nonlinear state estimation; however, its performance is limited by the static parameterization of the Unscented Transform (UT). Conventional weighting schemes, governed by fixed scaling parameters, assume implicit Gaussianity and fail to adapt to time-varying dynamics or heavy-tailed measurement noise. This work introduces the Meta-Adaptive UKF (MA-UKF), a framework that reformulates sigma-point weight synthesis as a hyperparameter optimization problem addressed via memory-augmented meta-learning. Unlike standard adaptive filters that rely on instantaneous heuristic corrections, our approach employs a Recurrent Context Encoder to compress the history of measurement innovations into a compact latent embedding. This embedding informs a policy network that dynamically synthesizes the mean and covariance weights of the sigma points at each time step, effectively governing the filter's trust in the prediction versus the measurement. By optimizing the system end-to-end through the filter's recursive logic, the MA-UKF learns to maximize tracking accuracy while maintaining estimation consistency. Numerical benchmarks on maneuvering targets demonstrate that the MA-UKF significantly outperforms standard baselines, exhibiting superior robustness to non-Gaussian glint noise and effective generalization to out-of-distribution (OOD) dynamic regimes unseen during training.
\end{abstract}


\section{Introduction}
\label{Introduction}

The Unscented Kalman Filter (UKF) has established itself as a standard of nonlinear state estimation since its inception by Julier and Uhlmann~\cite{julier1997new}. By propagating a deterministic set of sigma points through the nonlinear system dynamics, the UKF captures posterior moments accurate to the third order for Gaussian inputs, circumventing the linearization errors inherent in the Extended Kalman Filter (EKF) without requiring explicit Jacobian calculations~\cite{wan2000unscented}. However, the performance of the standard UKF is strictly governed by the parameterization of the Unscented Transform (UT), specifically, the scaling parameters ($\alpha$, $\beta$, $\kappa$) that determine the spread and weighting of the sigma points. While early efforts have explored model-based learning to optimize these sigma points~\cite{turner2012model}, conventional implementations still largely rely on static parameters selected \textit{a priori} based on implicit assumptions of Gaussianity and stationary noise statistics. Consequently, the filter's adaptability is severely limited in complex environments characterized by maneuvering targets, non-stationary dynamics, or heavy-tailed measurement noise~\cite{bar2001estimation}.

To address these limitations, the field of Adaptive Kalman Filtering (AKF) has historically relied on heuristic noise estimation methods, such as Sage-Husa estimators~\cite{sage1969adaptive} or Interacting Multiple Model (IMM) architectures~\cite{blom1988interacting}. While IMMs offer robustness by switching between varying dynamic models, they are computationally expensive and rely on a pre-defined, finite set of modes. More recently, the convergence of signal processing and deep learning has given rise to the broader paradigm of AI-aided Kalman filters~\cite{shlezinger2025artificial}. Historically, the relationship between recursive estimators and neural networks was often inverted, such as using EKFs to explicitly train the weights of Recurrent Neural Networks (RNNs)~\cite{wang2011convergence}. Today, deep learning actively enhances state estimation logic. Recent literature features model-based deep learning for maneuvering targets~\cite{forti2023model}, self-supervised deep state-space models for tracking~\cite{wang2025self}, and hybrid architectures that fuse RNNs with Kalman variants for complex trajectory prediction~\cite{jia2024multiple} and real-time sensor drift compensation~\cite{li2021recurrent}.

In parallel, advancements in Differentiable Filtering enable filter parameters to be optimized end-to-end via backpropagation~\cite{haarnoja2016backprop,jonschkowski2018differentiable}. This concept has successfully been expanded across various Bayesian estimators, including end-to-end learning for differentiable particle filters~\cite{wen2021end} and neural-augmented adaptive grids for point-mass filters~\cite{trejbal2025neural}. For continuous state spaces, approaches such as KalmanNet~\cite{revach2022kalmannet} learn to dynamically adjust the Kalman gain or noise covariances from data streams. While these methods improve state estimation, many largely treat the underlying sigma-point geometry as a fixed inductive bias.

Recognizing this gap, modern studies have begun to explicitly target the adaptation of the Unscented Transform. Recent frameworks have proposed data-driven unknown input estimation for sigma-point filters~\cite{loo2024sigma}, and utilized RNNs to auto-tune standard UKF hyperparameters dynamically~\cite{fan2024rnn}. Furthermore, neural networks have recently been proposed to fine-tune the explicit spatial placement of sample points in Gaussian filters~\cite{liu2025finetuning}. However, these approaches still largely focus on tuning standard scalar constraints or rely on instantaneous heuristic corrections, stopping short of reformulating the full synthesis of the UT as a continuous hyper-control problem.

Leveraging advancements in Gradient-Based Meta-Learning~\cite{duan2016rl,finn2017model}, particularly foundational works that cast "learning to learn" as a temporal sequence modeling problem via Memory-Augmented architectures~\cite{wang2016learning, mishra2017simple}, we propose a novel framework that reformulates the tuning of sigma-point weights as a sequential decision-making process. In this context, the filter is viewed as a differentiable computational graph where hyperparameters are adapted in real-time to unseen dynamic regimes. This represents a shift from static model-based estimation to context-aware meta-learning, where adaptation logic is amortized into the hidden states of an RNN.

In this work, we introduce the Meta-Adaptive UKF (MA-UKF). In contrast to end-to-end neural architectures that bypass state-space modeling, our framework maintains the structural integrity of the Bayesian recursion while adaptively regulating the UT parameters through a learned policy. Central to our architecture is a Recurrent Context Encoder~\cite{cho2014learning}, which compresses the history of measurement innovations into a latent embedding. This embedding enables the system to distinguish between dynamic maneuvers and sensor anomalies without explicit mode-switching logic. A policy network then maps this context to the optimal mean and covariance weights for the sigma points, satisfying the mathematical convexity constraints of the UT. By training the system end-to-end using analytical gradients through time~\cite{franceschi2017forward}, the MA-UKF learns to balance instantaneous tracking accuracy with long-term estimation stability.

The contributions of this paper are threefold:
\begin{enumerate}
   \item \textbf{Differentiable Meta-Filtering:} We cast the Unscented Transform parameterization as a bi-level optimization problem within a differentiable computational graph. This enables end-to-end learning of optimal, data-driven sigma-point weights via analytical gradients through time..
   
   \item \textbf{Memory-Augmented Adaptation:} We introduce a Recurrent Context Encoder that compresses innovation history into a latent embedding. This temporal context drives a learned policy to dynamically modulate the sigma-point weights in real-time, effectively distinguishing actual maneuvers from sensor anomalies.
   
   \item \textbf{Robustness and OOD Generalization:} We demonstrate that the MA-UKF significantly outperforms optimized standard filters and IMM baselines under heavy-tailed glint noise. Crucially, the learned policy exhibits strong out-of-distribution (OOD) generalization, maintaining accurate tracking against high-agility maneuvers unseen during training.
\end{enumerate}

The remainder of this paper is organized as follows: Section~\ref{sec:theoretical} provides the theoretical framework, detailing the Unscented Transform and the Meta-Learning objective. Section~\ref{sec:maukf} describes the proposed MA-UKF architecture and the training methodology. Section~\ref{sec:experiment} presents the experimental setup and comparative results. Finally, Section~\ref{sec:conclusions} concludes the paper and outlines future research directions.

\section{Theoretical Framework}\label{sec:theoretical}

\subsection{Problem Definition}
Consider the state estimation problem for a discrete-time nonlinear dynamic system. While standard filtering approaches assume the system follows nominal Gaussian statistics, practical applications often involve regime shifts and distributional mismatches. Let the true generative process be defined by:

\begin{align}
    \mathbf{x}_{k} &= f(\mathbf{x}_{k-1}) + \mathbf{w}_k, \quad \mathbf{w}_k \sim \mathcal{D}_w(\cdot), \\
    \mathbf{z}_{k}   &= h(\mathbf{x}_k) + \mathbf{v}_k, \quad \mathbf{v}_k \sim \mathcal{D}_v(\cdot),
\end{align}

\noindent where $\mathbf{x}_k \in \mathbb{R}^{n_x}$ is the state vector  at time step $k$, $\mathbf{z}_k \in \mathbb{R}^{n_z}$ is the measurement vector, and $f : \mathbb{R}^{n_x}\rightarrow\mathbb{R}^{n_x}$ and $h : \mathbb{R}^{n_x}\rightarrow\mathbb{R}^{n_z}$ are nonlinear transition and observation functions, respectively.

In standard formulations, the process noise $\mathbf{w}_k$ and measurement noise $\mathbf{v}_k$ are assumed to be drawn from fixed Gaussian distributions, i.e., $\mathcal{D}_w = \mathcal{N}(\mathbf{0}, \mathbf{Q})$ and $\mathcal{D}_v = \mathcal{N}(\mathbf{0}, \mathbf{R})$. However, in robust tracking scenarios (e.g., maneuvering targets or radar glint), the true distributions $\mathcal{D}_w$ and $\mathcal{D}_v$ are time-varying, multimodal, or characterized by non-zero skewness and high kurtosis.

Our objective is to derive an optimal estimator $\mathcal{F}_{\theta}$ parameterized by $\theta$, which produces a state estimate $\hat{\mathbf{x}}_k$ at each time step. We seek to minimize the expected cumulative estimation error over a finite horizon $T$:

\begin{equation}
    \theta^* = \arg \min_{\theta} \mathbb{E}_{\tau \sim \mathcal{P}_\tau} \left[ \sum_{k=1}^{T} \|\mathbf{x}_k - \mathcal{F}_\theta(\hat{\mathbf{x}}_{k-1}, \mathbf{z}_k)\|_2^2 \right],
\end{equation}

\noindent where $\tau = (\mathbf{x}_{0:T}, \mathbf{z}_{1:T})$ represents a trajectory sampled from the true system distribution $\mathcal{P}_\tau$. Unlike traditional adaptive filtering, which attempts to explicitly estimate time-varying noise covariances $\mathbf{Q}$ or $\mathbf{R}$, we seek to optimize the internal integration parameters of the filter itself directly via the parameter set $\theta$.

\subsection{Standard Unscented Kalman Filter Formulation}
The UKF approximates the state distribution using a set of $2n_x+1$ sigma points, $\mathcal{X}_{k-1} = \{\mathcal{X}_{k-1}^i\}_{i=0}^{2n_x}$, with associated weights $W^{(m)}_i$ for the mean and $W^{(c)}_i$ for the covariance. Given the posterior estimate $\hat{\mathbf{x}}_{k-1}$ and covariance $\mathbf{P}_{k-1}$ from the previous step, the sigma points are generated as:

\begin{align}
      \mathcal{X}_{k-1}^0 &= \hat{\mathbf{x}}_{k-1} \label{eq:sigma_0} \\ 
      \mathcal{X}_{k-1}^i &= \hat{\mathbf{x}}_{k-1} + \left[\sqrt{(n_x+\lambda)\mathbf{P}_{k-1}}\right]_i, \quad i = 1, \dots, n_x \label{eq:sigma_plus} \\
      \mathcal{X}_{k-1}^{i+n_x} &= \hat{\mathbf{x}}_{k-1} - \left[\sqrt{(n_x+\lambda)\mathbf{P}_{k-1}}\right]_i, \quad i = 1, \dots, n_x \label{eq:sigma_minus}
\end{align}

\noindent where $[\cdot]_i$ denotes the $i$-th column of the matrix square root, typically computed via the lower-triangular Cholesky decomposition. The scaling parameter is defined as $\lambda = \alpha^2(n_x+\kappa)-n_x$, where $\alpha$ controls the spread of the sigma points and $\kappa$ is a secondary scaling parameter.

The constant weights corresponding to these points are given by:
\begin{align} 
    W_0^{(m)} &= \frac{\lambda}{n_x+\lambda}, \label{eq:wm0}\\ 
    W_0^{(c)} &= \frac{\lambda}{n_x+\lambda} + (1-\alpha^2+\beta), \label{eq:wc0}\\ 
    W_i^{(m)} &= W_i^{(c)} = \frac{1}{2(n_x+\lambda)}, \quad i = 1 \dots 2n_x. \label{eq:wmwc}
\end{align}
where $\beta$ is a scalar parameter used to incorporate prior knowledge of the underlying distribution (typically $\beta=2$ for optimal Gaussian approximations).

\noindent \textbf{Prediction Step:} The sigma points are propagated through the nonlinear process function:
\begin{equation} 
    \mathcal{X}_{k|k-1}^i = f(\mathcal{X}_{k-1}^i). \label{eq:sigma_propagate}
\end{equation}
The \textit{a priori} state mean and covariance are computed via weighted recombination:
\begin{align} 
    \hat{\mathbf{x}}^{-}_{k} &= \sum_{i=0}^{2n_x} W_i^{(m)} \mathcal{X}_{k|k-1}^i, \label{eq:x_minus_recombination} \\ 
    \mathbf{P}^{-}_{k} &= \sum_{i=0}^{2n_x} W_i^{(c)} (\mathcal{X}_{k|k-1}^i - \hat{\mathbf{x}}^{-}_{k})(\mathcal{X}_{k|k-1}^i - \hat{\mathbf{x}}^{-}_{k})^\top + \mathbf{Q}_k.\label{eq:p_minus_recombination} 
\end{align}

\noindent \textbf{Update Step:} The propagated sigma points are then passed through the measurement function:
\begin{equation} 
   \mathcal{Z}_{k|k-1}^i = h(\mathcal{X}_{k|k-1}^i). 
\end{equation}
The predicted measurement mean $\hat{\mathbf{z}}^{-}_{k}$, innovation covariance $\mathbf{P}_{zz}$, and cross-covariance $\mathbf{P}_{xz}$ are computed as:
\begin{align} 
    \hat{\mathbf{z}}^{-}_{k} &= \sum_{i=0}^{2n_x} W_i^{(m)} \mathcal{Z}_{k|k-1}^i, \\ 
    \mathbf{P}_{zz} &= \sum_{i=0}^{2n_x} W_i^{(c)} (\mathcal{Z}_{k|k-1}^i - \hat{\mathbf{z}}^{-}_{k})(\mathcal{Z}_{k|k-1}^i - \hat{\mathbf{z}}^{-}_{k})^\top + \mathbf{R}_{k}, \\
   \mathbf{P}_{xz} &= \sum_{i=0}^{2n_x} W_i^{(c)} (\mathcal{X}_{k|k-1}^i - \hat{\mathbf{x}}^{-}_{k})(\mathcal{Z}_{k|k-1}^i - \hat{\mathbf{z}}^{-}_{k})^\top. 
\end{align}

\noindent Finally, the Kalman gain $\mathbf{K}_k$ is computed, and the posterior estimates are updated:
\begin{align} 
    \mathbf{K}_{k} &= \mathbf{P}_{xz} \mathbf{P}_{zz}^{-1}, \label{eq:kcompute}\\
    \hat{\mathbf{x}}_{k} &= \hat{\mathbf{x}}^{-}_{k} + \mathbf{K}_{k} (\mathbf{z}_{k} - \hat{\mathbf{z}}^{-}_{k}), \label{eq:xupdate}\\
    \mathbf{P}_{k} &= \mathbf{P}^{-}_{k} - \mathbf{K}_{k} \mathbf{P}_{zz} \mathbf{K}_{k}^\top. \label{eq:pupdate}
\end{align}

\subsection{The Paradigm of Differentiable Filtering}
Differentiable Filtering represents the convergence of Bayesian recursive estimation and deep learning. Traditionally, Kalman filters and neural networks were treated as separate entities: the former as model-based estimators and the latter as data-driven approximators. In the Differentiable Filtering paradigm, the filter itself is cast as a computational graph, analogous to a RNN, but with an inductive bias strictly imposed by the physical equations of motion and the Bayesian update rules~\cite{haarnoja2016backprop,jonschkowski2018differentiable,kloss2020accurate}.

Let $\mathcal{F}_{\theta}$ denote the single-step UKF update function defined by equations~\eqref{eq:sigma_0} through~\eqref{eq:pupdate}. We can view the filter state at time $k$ as a tuple $\mathbf{s}_k = (\hat{\mathbf{x}}_k, \mathbf{P}_k)$. The filter evolution is then expressed as a differentiable recurrence:
\begin{equation}
    \mathbf{s}_k = \mathcal{F}_{\theta}(\mathbf{s}_{k-1}, \mathbf{z}_k),
\end{equation}

\noindent where $\theta$ represents the tunable parameters (e.g., the policy network weights governing sigma-point scaling). Crucially, operations such as the Cholesky decomposition required for the matrix square root in~\eqref{eq:sigma_plus} and~\eqref{eq:sigma_minus} are differentiable for positive-definite matrices. This enables the continuous flow of gradients through the stochastic covariance parameters of the filter~\cite{murray2016differentiation}.

To optimize $\theta$, we compute the gradient of the cumulative loss $\mathcal{L}$ with respect to the parameters. Since the parameters $\theta$ influence the entire trajectory, we employ Backpropagation Through Time (BPTT)~\cite{werbos2002backpropagation}. Applying the chain rule, the total gradient is the accumulation of gradients at each time step:
\begin{equation}
    \frac{d \mathcal{L}}{d \theta} = \sum_{k=1}^T \frac{\partial \mathcal{L}_k}{\partial \mathbf{s}_k} \frac{d \mathbf{s}_k}{d \theta},
\end{equation}
where the total state derivative $\frac{d \mathbf{s}_k}{d \theta}$ is computed recursively:
\begin{equation}
    \frac{d \mathbf{s}_k}{d \theta} = \frac{\partial \mathcal{F}_{\theta}}{\partial \theta} + \frac{\partial \mathcal{F}_{\theta}}{\partial \mathbf{s}_{k-1}} \cdot \frac{d \mathbf{s}_{k-1}}{d \theta}.
\end{equation}
Here, $\frac{\partial \mathcal{F}_{\theta}}{\partial \mathbf{s}_{k-1}}$ is the Jacobian of the UKF update step, capturing exactly how uncertainty $\mathbf{P}_{k-1}$ propagates through the system nonlinearities $k$. By unrolling this computational graph and optimizing $\theta$ via automatic differentiation, we successfully retain the structural interpretability of the Kalman filter while gaining the data-driven adaptability of neural networks.

\subsection{Context-Driven Meta-Policy Formulation}

A fundamental limitation of the standard UKF is the static parameterization of the UT. The weights $\mathcal{W} = \{W^{(m)}, W^{(c)}\}$ enforce a time-invariant geometry on the sigma points, implicitly assuming that the system's nonlinearity and noise statistics remain stationary. In dynamic environments, this rigidity forces a compromise between tracking agility (high reactivity) and noise suppression (high stability).

To resolve this, we reformulate the filtering problem within the paradigm of \textit{Differentiable Programming}~\cite{mishra2017simple}. We treat the UKF not as a fixed algorithm, but as a directed computational graph, parameterized by a meta-policy $\pi_{\theta}$. This policy functions as a hyper-controller, dynamically modulating the filter's integration parameters based on the estimation context.

At each time step $k$, the policy processes the measurement innovation $\boldsymbol{\nu}_k$. This proxy signal is used to update a recurrent hidden state $\mathbf{h}_k$. This latent state summarizes the history of the trajectory, allowing the network to distinguish between transient maneuvers and non-physical sensor noise. The adaptive weights are then synthesized as:
\begin{align}
    \mathbf{h}_k &= \text{RNN}_{\theta}(\boldsymbol{\nu}_k, \mathbf{h}_{k-1}), \\
    \mathcal{W}_k &= \{W^{(m)}_k, W^{(c)}_k\} = \pi_{\theta}(\mathbf{h}_k).
\end{align}
These dynamic weights are immediately injected into the standard UKF prediction and update cycle to produce the posterior state estimate $\hat{\mathbf{x}}_k$.

This architecture constitutes a bi-level optimization problem~\cite{hospedales2021meta}. The \textit{inner loop} performs the recursive Bayesian state estimation, while the \textit{outer loop} optimizes the policy parameters $\theta$. Since the constituent operations of the UKF, including matrix multiplication, Cholesky decomposition, and linear system solution, are differentiable almost everywhere, the gradient of the cumulative estimation error $\nabla_{\theta} \mathcal{L}$ can be computed analytically via BPTT.

\section{Meta-Adaptive Unscented Kalman Filter}\label{sec:maukf}
\begin{figure*}[htbp]
    \centering
    \includegraphics[width=1.0
    \linewidth]{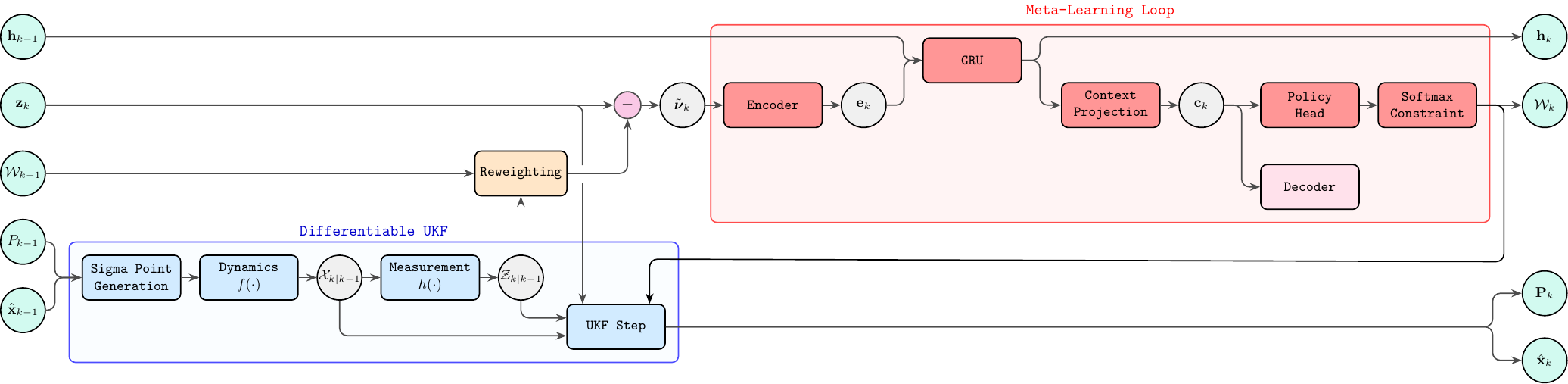}
    \caption{Architecture of the Meta-Adaptive UKF (MA-UKF). A Recurrent Context Encoder processes the innovation history to generate a latent context embedding, which drives a policy network to modulate the sigma-point weights in real-time.}
    \label{fig:architecture}
\end{figure*}

This section details the architecture of the proposed Meta-Adaptive UKF (MA-UKF). The framework is designed to decouple the physical state propagation from the statistical parameterization of the filter. As illustrated in Fig.~\ref{fig:architecture}, the system comprises three functional modules: innovation feature extraction, recurrent context encoding, and convex weight synthesis.

\subsection{Innovation Feature Extraction}
In recursive estimation, the innovation sequence $\boldsymbol{\nu}_k$ constitutes the primary signal for adaptation. While classical adaptive filters utilize scalar metrics such as the Normalized Innovation Squared (NIS) to trigger heuristic logic~\cite{8454982}, raw innovation magnitudes are highly sensitive to sensor modality, target range, and scale.

To ensure numerical stability and facilitate efficient gradient propagation, we employ a learnable feature extraction stage. To circumvent the cyclic dependency we compute a \textit{proxy innovation} $\tilde{\boldsymbol{\nu}}_k \in \mathbb{R}^{n_z}$  using the nominal geometric spread from the previous time step:

\begin{equation}
    \tilde{\boldsymbol{\nu}}_k = \mathbf{z}_k - \sum_{i=0}^{2n_x} W_{i, k-1}^{(m)} \mathcal{Z}_{k|k-1}^i.
\end{equation}

This raw signal is projected into a high-dimensional feature space via a linear projection followed by Layer Normalization~\cite{ba2016layernormalization}:

\begin{equation}
    \mathbf{e}_k = \text{ReLU}(\text{LayerNorm}(\mathbf{W}_{\text{in}} \tilde{\boldsymbol{\nu}}_k + \mathbf{b}_{\text{in}})),
\end{equation}
where $\mathbf{W}_{\text{in}} \in \mathbb{R}^{d_{h} \times n_z}$ is a learnable projection matrix, $\mathbf{b}_{\text{in}} \in \mathbb{R}^{d_{h}}$ is a bias, and $\mathbf{e}_k \in \mathbb{R}^{d_{h}}$ denotes the static feature vector of the instantaneous residual. This normalization is essential for robustness in the glint noise regime, where outlier measurements may exceed nominal variances by several orders of magnitude.

\subsection{Recurrent Context Encoding}
Instantaneous residuals alone are insufficient to distinguish between measurement outliers and the onset of a maneuver. While both phenomena yield large norms $\|\boldsymbol{\nu}_k\|$, they exhibit distinct temporal structures: glint is typically uncorrelated and transient, whereas maneuvers manifest as correlated, low-frequency trends.

To capture these temporal dependencies, we employ a Gated Recurrent Unit (GRU). The GRU functions as a non-linear Infinite Impulse Response (IIR) filter, integrating information over time to update a latent context embedding $\mathbf{h}_k \in \mathbb{R}^{d_{h}}$. The update dynamics are governed by:
\begin{align}
    \mathbf{u}_k &= \sigma_g(\mathbf{W}_u \mathbf{e}_k + \mathbf{U}_u \mathbf{h}_{k-1} + \mathbf{b}_u), \\
    \mathbf{r}_k &= \sigma_g(\mathbf{W}_r \mathbf{e}_k + \mathbf{U}_r \mathbf{h}_{k-1} + \mathbf{b}_r), \\
    \tilde{\mathbf{h}}_k &= \tanh(\mathbf{W}_h \mathbf{e}_k + \mathbf{U}_h (\mathbf{r}_k \odot \mathbf{h}_{k-1}) + \mathbf{b}_h), \\
    \mathbf{h}_k &= (1 - \mathbf{u}_k) \odot \mathbf{h}_{k-1} + \mathbf{u}_k \odot \tilde{\mathbf{h}}_k,
\end{align}
where $\sigma_g(\cdot)$ denotes the sigmoid activation, $\odot$ represents the Hadamard product, and $\mathbf{u}_k$ and $\mathbf{r}_k$ are the update and reset gates, respectively. The parameter matrices are defined as $\mathbf{W}_u, \mathbf{W}_r, \mathbf{W}_h \in \mathbb{R}^{d_h \times d_h}$ for the input weights, and $\mathbf{U}_u, \mathbf{U}_r, \mathbf{U}_h \in \mathbb{R}^{d_h \times d_h}$ for the recurrent weights, with corresponding biases $\mathbf{b}_u, \mathbf{b}_r, \mathbf{b}_h \in \mathbb{R}^{d_h}$. This mechanism allows the policy to selectively suppress high-frequency noise while retaining sensitivity to genuine dynamic shifts.

\subsection{Convex Sigma-Point Weight Synthesis}
The core contribution of the MA-UKF is the learned policy mapping $\pi_{\theta}: \mathbf{h}_k \rightarrow \mathcal{W}_k$, which dynamically synthesizes the Unscented Transform parameters. First, the high-dimensional hidden state is projected onto a lower-dimensional context manifold $\mathbf{c}_k \in \mathbb{R}^{d_{p}}$:
\begin{equation}
    \mathbf{c}_k = \text{ReLU}(\text{LayerNorm}(\mathbf{W}_{proj} \mathbf{h}_k + \mathbf{b}_{proj})),
\end{equation}
where $\mathbf{W}_{proj} \in \mathbb{R}^{d_p \times d_h}$ and $\mathbf{b}_{proj} \in \mathbb{R}^{d_p}$.

 A critical constraint in Unscented filtering is the numerical stability of the covariance update, which requires the covariance matrices to remain positive semi-definite. Standard parameterizations (e.g., utilizing negative $\lambda$ values) can yield non-positive definite matrices, inducing numerical instability and potentially leading to Cholesky decomposition failures during training. To guarantee stability, we enforce a \textit{convexity constraint} on the sigma-point weights. The mean weights $W^{(m)}$ and covariance weights $W^{(c)}$ are generated via a Softmax function:
 \begin{align}
     \mathbf{l}^{(m)}_k &= \mathbf{W}_{\pi}^{(m)} \mathbf{c}_k + \mathbf{b}_{\pi}^{(m)}, \\
     W_{i,k}^{(m)} &= \frac{\exp(l_{i,k}^{(m)})}{\sum_{j=0}^{2n_x} \exp(l_{j,k}^{(m)})},
 \end{align}
where $\mathbf{W}_{\pi}^{(m)} \in \mathbb{R}^{(2n_x+1) \times d_p}$ and $\mathbf{b}_{\pi}^{(m)} \in \mathbb{R}^{2n_x+1}$. The covariance weights $W^{(c)}$ are computed similarly. This constraint ensures that $\sum_i W_i = 1$ and $W_i > 0$, guaranteeing that the predicted covariance $\mathbf{P}_k^-$ remains positive definite. While this restriction prevents the filter from capturing the higher-order kurtosis effects possible with negative weights, it provides the strict stability guarantee required for end-to-end differentiable training, ensuring that the Cholesky decompositions in the UKF remain valid throughout the optimization process.

\subsection{Algorithm and Computational Flow}
The recursive execution of the MA-UKF is summarized in Algorithm~\ref{alg:ma_ukf}. A key design principle is the decoupling of the optimization burden: the computationally expensive training of the meta-policy is performed offline, allowing the dynamic, real-time adaptation of the filter parameters to be executed online via a highly efficient single forward pass of the lightweight RNN.

\begin{algorithm}[H]
\caption{Meta-Adaptive UKF Recursive Step}
\label{alg:ma_ukf}
\begin{algorithmic}[1]
\REQUIRE $\hat{\mathbf{x}}_{k-1}, \mathbf{P}_{k-1}$, $\mathcal{W}_{k-1}$, $\mathbf{h}_{k-1}$, $\mathbf{z}_k$, $\mathbf{Q}_k, \mathbf{R}_k$, $\gamma$

\vspace{0.2cm}

\STATE $\mathcal{X}_{k-1} \gets \left[ \hat{\mathbf{x}}_{k-1}, \hat{\mathbf{x}}_{k-1} \pm \sqrt{\gamma \mathbf{P}_{k-1}} \right]$

\vspace{0.2cm}

\STATE $\mathcal{X}_{k|k-1} \gets f(\mathcal{X}_{k-1})$

\STATE $\mathcal{Z}_{k|k-1} \gets h(\mathcal{X}_{k|k-1})$

\STATE $\tilde{\boldsymbol{\nu}}_{k} \leftarrow \mathbf{z}_k - \sum_{i=0}^{2n_x} W_{i, k-1}^{(m)} \mathcal{Z}_{k|k-1}$

\vspace{0.2cm}

\STATE $\mathbf{e}_k \gets \mathrm{ReLU}(\mathrm{LayerNorm}(\mathbf{W}_{in}\tilde{\boldsymbol{\nu}}_{k} + \mathbf{b}_{in}))$

\STATE $\mathbf{h}_k \gets \mathrm{GRU} (\mathbf{e}_k, \mathbf{h}_{k-1})$

\STATE $\mathcal{W}_{k} = \{W_k^{(m)}, W_k^{(c)}\} \gets \pi_{\theta}(\mathbf{h}_k)$

\vspace{0.2cm}

\STATE $\hat{\mathbf{x}}_k^- \gets \sum_{i=0}^{2n_x} W_{i,k}^{(m)} \mathcal{X}^{i}_{k|k-1}$

\STATE $\mathbf{P}_k^- \gets \sum_{i=0}^{2n_x} W_{i,k}^{(c)} ( \mathcal{X}^{i}_{k|k-1} - \hat{\mathbf{x}}_k^- ) ( \mathcal{X}^{i}_{k|k-1} - \hat{\mathbf{x}}_k^- )^\top + \mathbf{Q}_k$

\vspace{0.2cm}

\STATE $\hat{\mathbf{z}}_k^- \gets \sum_{i=0}^{2n_x} W_{i,k}^{(m)} \mathcal{Z}^{i}_{k|k-1}$

\STATE $\mathbf{P}_{zz} \gets \sum_{i=0}^{2n_x} W_{i,k}^{(c)} ( \mathcal{Z}^{i}_{k|k-1} - \hat{\mathbf{z}}_k^-)( \mathcal{Z}^{i}_{k|k-1} - \hat{\mathbf{z}}_k^-)^\top + \mathbf{R}_k$

\STATE $\mathbf{P}_{xz} \leftarrow \sum_{i=0}^{2n_x} W_{i,k}^{(c)} ( \mathcal{X}^{i}_{k|k-1} - \hat{\mathbf{x}}_k^- ) ( \mathcal{Z}^{i}_{k|k-1} - \hat{\mathbf{z}}_k^- )^\top$

\vspace{0.2cm}

\STATE $\mathbf{K}_k \leftarrow \mathbf{P}_{xz} \mathbf{P}_{zz}^{-1}$
\STATE $\hat{\mathbf{x}}_k \leftarrow \hat{\mathbf{x}}_k^- + \mathbf{K}_k (\mathbf{z}_k - \hat{\mathbf{z}}_k^-)$
\STATE $\mathbf{P}_k \leftarrow \mathbf{P}_k^- - \mathbf{K}_k \mathbf{P}_{zz} \mathbf{K}_k^\top$

\vspace{0.2cm}

\RETURN $\hat{\mathbf{x}}_k, \mathbf{P}_k, \mathcal{W}_{k}, \mathbf{h}_k$
\end{algorithmic}
\end{algorithm}

\subsection{Computational Complexity Analysis}

For real-time tracking applications, computational efficiency is paramount. The overhead introduced by the Meta-Adaptive block is governed by the feature extraction, GRU state update, and policy projection layers, yielding an additive time complexity of $\mathcal{O}(d_{h}^2 + d_h n_z + d_h d_p + n_x d_p)$. Given the compact dimensionality of the proposed architecture ($n_x=5, n_z=2, d_{h}=32, d_p=16$), this neural inference sequence strictly requires on the order of a few thousand floating-point operations (FLOPs) per recursive cycle. By leveraging highly optimized dense matrix operations, this forward pass incurs sub-microsecond latency on contemporary microprocessors. Consequently, the total execution time of the MA-UKF imposes negligible overhead relative to a standard single-model UKF, whose performance bottleneck remains the $\mathcal{O}(n_x^3)$ Cholesky decomposition and covariance updates. Crucially, this allows the MA-UKF to achieve robust adaptation while entirely circumventing the $\mathcal{O}(M \cdot n_x^3)$ multiplicative scaling inherent to IMM architectures evaluating $M$ parallel hypotheses.

\subsection{Training via End-to-End Analytical Gradients}

We formulate the policy optimization as a supervised regression problem over a dataset of $N$ trajectories. The loss function $\mathcal{L}_{\theta}$ comprises a primary tracking error term and an auxiliary regularization term:
\begin{equation}
\mathcal{L}_{\theta} = \sum_{k=1}^T \|\mathbf{x}_k - \mathcal{F}_\theta(\hat{\mathbf{x}}_{k-1}, \mathbf{z}_k)\|_2^2 + \lambda_{aux} \|\tilde{\boldsymbol{\nu}}_k - g(\mathbf{c}_k)\|_2^2.
\end{equation}
Here $g(\cdot)$ is an auxiliary decoder, weighted by $\lambda_{aux}$, that reconstructs the innovation from the latent context $\mathbf{c}_k$. This regularization encourages the latent state $\mathbf{h}_k$ to retain a rich representation of the innovation sequence, preventing mode collapse during training.

\section{Experimental Evaluation}\label{sec:experiment}

To validate the efficacy of the proposed MA-UKF, we conducted a series of numerical experiments focusing on robust tracking in the presence of heavy-tailed measurement noise and unmodeled dynamic maneuvers. We benchmark the method against a nominal UKF, a hyperparameter-optimized UKF, and an IMM filter.

\subsection{Simulation Environment and Datasets}

We consider a 2D radar tracking scenario where the state vector $\mathbf{x} = [p_x, v_x, p_y, v_y, \omega]^\top$ evolves according to the Coordinated Turn (CT) kinematic model with a discrete-time sampling interval of $\Delta t = 0.1$ seconds. The radar provides noisy range and bearing measurements:
\begin{equation}
    \mathbf{z}_k = \begin{bmatrix} \sqrt{p_x^2 + p_y^2},\ \arctan(p_y/p_x) \end{bmatrix} + \mathbf{v}_k.
\end{equation}

\subsubsection{Training Regime (Stochastic CT)}
The MA-UKF is trained on a synthetic dataset of stochastic trajectories generated via the standard CT model. For each trajectory, the initial state $\mathbf{x}_0$ is randomized: positions $p_x, p_y$ are sampled from $\mathcal{U}(-1000, 1000)$ m, while the velocity is defined by a speed $v \sim \mathcal{U}(10, 30)$ m/s and a heading $\varphi \sim \mathcal{U}(0, 2\pi)$ rad. The turn rate is drawn from $\omega \sim \mathcal{U}([-0.5, -0.1] \cup [0.1, 0.5])$ rad/s, ensuring the model is trained on distinct left and right maneuvers rather than near-zero turn rates. The noise $\mathbf{v}_k$ is drawn from a Gaussian Mixture Model (GMM):

\begin{equation}
    \mathbf{v}_k \sim (1-\epsilon)\mathcal{N}(0, \mathbf{R}) + \epsilon\mathcal{N}(0, \eta \mathbf{R}),
\end{equation}
where $\epsilon=0.1$ represents a 10\% probability of a glint outlier (signal spike), and $\mathbf{R}$ is the nominal sensor noise covariance, and $\eta=20$ is the glint scaling factor.

\subsubsection{Evaluation Regime (OOD Generalization)}
Crucially, the evaluation is performed on a dataset representing a dynamic regime unseen during training. We utilize a high-agility weave maneuver, where the target undergoes sinusoidal acceleration with randomized parameters:

\begin{equation}
    \mathbf{a}(t) = \begin{bmatrix} A_x \sin(\omega_x t),\  A_y \cos(\omega_y t) \end{bmatrix}^\top.
\end{equation}

For each evaluation sequence, the amplitudes $A_{x}$, $A_{y}$ are sampled from $\mathcal{U}([-20,-10]\cup[10, 20])\,\text{m/s}^2$ and frequencies $\omega_{x}$, $\omega_{y}$ from $\mathcal{U}(-2, 2)\,\text{rad/s}$. 

This trajectory violates the constant turn-rate assumption inherent in the filter's state transition function ($f(\cdot)$). Furthermore, we test the robustness of the filter by doubling the severity of the measurement outliers during evaluation, setting the glint scaling factor to $\eta=40$. This combination of unmodeled dynamics and extreme noise strictly tests the generalization capability of the learned policy.

\subsection{Implementation and Baselines}

All models were implemented in \texttt{Python} using the \texttt{JAX} and \texttt{Equinox} libraries for differentiable programming. The MA-UKF was trained for 1800 epochs using the Adam optimizer ($lr=10^{-3}$) with a batch size of 64 and sequence length of 60 on an NVIDIA RTX 3060 GPU.

\textbf{Baseline 1: Nominal UKF.} We compare against a standard UKF tuned for nominal performance. While the UT often employs a small spread ($\alpha=10^{-3}$) to approximate local linearization, such settings are numerically unstable in high-noise regimes. Consequently, we adopt the parameterization proposed by~\cite{julier1997new}: $\alpha=1.0$, $\beta=2.0$, and $\kappa=3-n_x$. This setting places sigma points at $\pm\sqrt{3}\sigma$ from the mean, capturing the fourth-order moment of the Gaussian distribution and providing sufficient geometric spread. The process noise covariance $\mathbf{Q}$ and measurement noise $\mathbf{R}$ are fixed to the nominal ground-truth values.

\textbf{Baseline 2: Optimized UKF (UKF$^\star$).} To establish a rigorous baseline, we utilized the \texttt{Optuna} framework to perform a hyperparameter optimization search over 100 trials.  The resulting parameters ($\alpha \approx 17.26$, $\beta \approx 2.59$, $\kappa \approx 0.15$) represent the empirical upper bound of performance for a static UKF on this specific dataset.

\textbf{Baseline 3: IMM-UKF.}: We also employ an IMM filter mixing two UKF modes: a Constant Velocity (CV) mode for straight-line motion and a CT mode for maneuvers~\cite{cork2007sensor}. The transition probability matrix diagonal is fixed at $\Pi_{ii} = 0.95$. A hyperparameter-optimized version (IMM-UKF$^\star$) is also evaluated.

\textbf{Proposed MA-UKF.} The policy network utilizes a GRU with a hidden dimension of $d_{h}=32$. The context encoder projects the innovation proxy $\tilde{\boldsymbol{\nu}}_{k}$ into the hidden space, and a subsequent projection layer maps the GRU output to a lower-dimensional manifold ($d_{p}=16$) before the final policy head synthesizes the sigma-point weights. Throughout both the training and evaluation phases, the scaling parameter governing the sigma-point spread is fixed at $\gamma = 3$. 

\subsection{Quantitative Results}

We conducted a Monte Carlo analysis encompassing $N=1000$ independent tracking episodes across both the training dataset and the unseen high-agility weave dataset, with each episode evaluated over a sequence length of 60 time steps. Table~\ref{tab:results} summarizes the Average Root Mean Square Error (ARMSE) and the associated standard deviation for the position estimates.

\begin{table}[h]
\centering
\caption{Monte Carlo Performance Benchmark (1000 Runs)}
\label{tab:results}
\begin{tabular}{lcc}
\hline
\textbf{Method} & \textbf{Training Regime} & \textbf{Unseen Maneuver} \\
& ARMSE {(m)} & ARMSE {(m)} \\ \hline
UKF & $\armse{105.0}{129.6}$ &  $\armse{196.0}{229.9}$ \\ 
IMM-UKF & $\armse{86.4}{121.5}$ & $\armse{184.5}{175.8}$ \\ 
UKF$^\star$ & $\armse{17.8}{14.5}$ &  $\armse{49.7}{33.9}$ \\ 
IMM-UKF$^\star$ & $\armse{18.5}{17.4}$ & $\armse{58.0}{48.8}$ \\ 
\textbf{MA-UKF (Ours)} & $\armse{\mathbf{6.3}}{\mathbf{7.3}}$ &  $\armse{\mathbf{44.6}}{\mathbf{28.8}}$ \\ \hline
\end{tabular}
\end{table}

\subsubsection{Robustness to Heavy-Tailed Noise}
In the training regime, where the generative motion model closely aligns with the filter's internal dynamics, baseline performance is predominantly limited by sensor anomalies. The nominal UKF yields an ARMSE of $105.0$\,m, while the optimized UKF$^\star$ achieves $17.8$\,m. This degradation is directly attributable to the heavy-tailed measurement noise. The standard weighted averaging operation of the UT is highly sensitive to outliers; consequently, a single glint measurement significantly biases the posterior mean and violates the implicit Gaussian assumption.

In contrast, the MA-UKF achieves an ARMSE of just $6.3$\,m, representing a 64.6\% reduction in error compared to the optimized UKF$^\star$ and a 94.0\% reduction against the nominal UKF. By analyzing the latent innovation embedding, the network identifies the spectral signature of glint, characterized by high amplitude and low temporal correlation. The system responds by dynamically down-weighting the sigma points. This mechanism seamlessly performs "soft" outlier rejection without the need for brittle validation gating or explicit thresholds.

\subsubsection{Generalization to Unseen Maneuvers (OOD)}
The evaluation regime leverages a high-agility weave trajectory that presents a severe structural mismatch to the filter's constant turn-rate prediction model. This scenario tests the algorithm's capacity to balance model trust against measurement reliance.

\begin{figure}[htbp]
    \centering
    \includegraphics[width=1.0\linewidth]{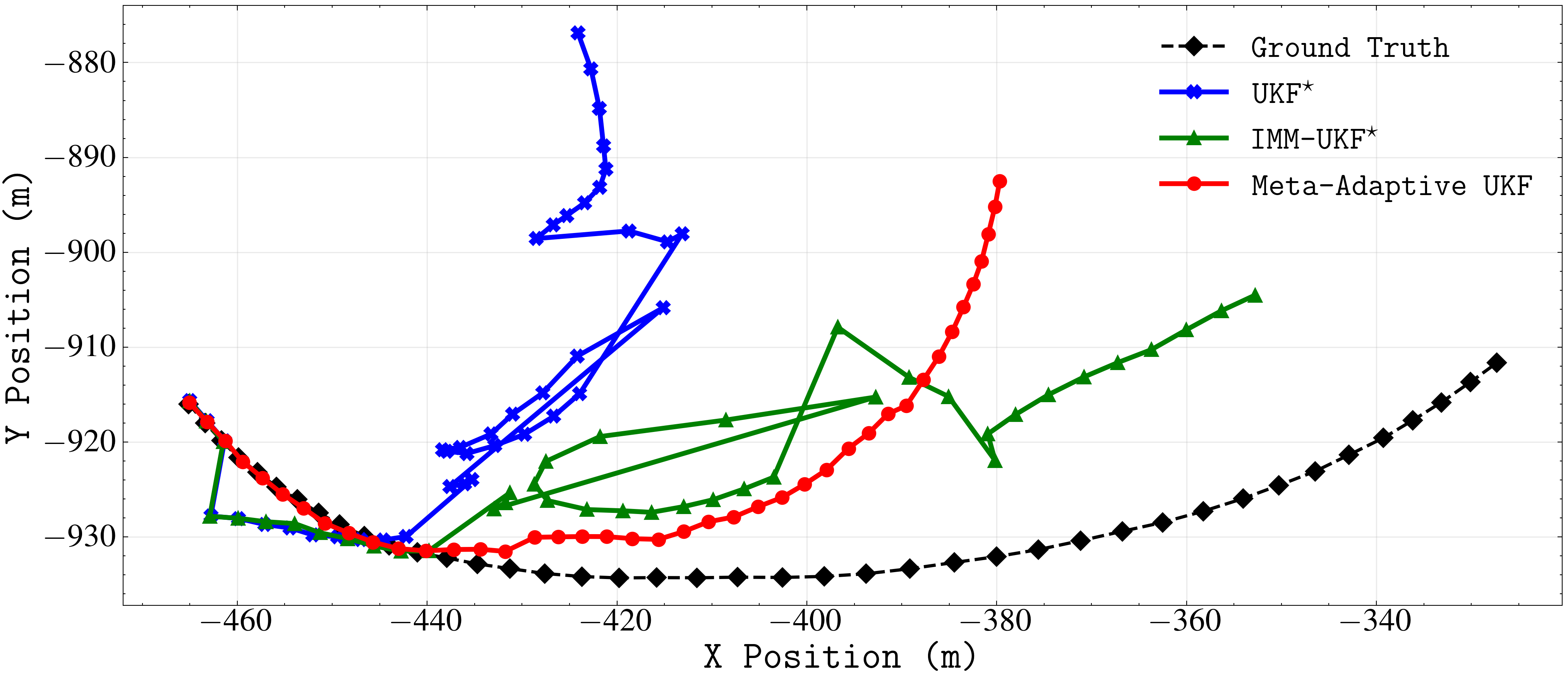} 
    \caption{Trajectory tracking performance on the OOD maneuver. The optimized UKF$^{\star}$ (Blue) suffers catastrophic divergence, while the IMM-UKF$^{\star}$ (Green) exhibits violent corrections. The MA-UKF (Red) sustains robust track continuity for longer,  and closely hugs the underlying ground truth despite the simultaneous model mismatch and severe glint noise.}
    \label{fig:trajectory}
\end{figure}

As illustrated in Fig.~\ref{fig:trajectory}, the baseline filters fail to handle the compounding severity of unmodeled dynamics and extreme noise. The UKF$^{\star}$ exhibits track divergence, losing the target entirely. The IMM-UKF$^{\star}$, while attempting to switch dynamic models to recapture the target, suffers from massive correction artifacts and erratic, high-amplitude jumps.

The MA-UKF, conversely, successfully preserves the structural geometry of the ground truth. It maintains consistent track continuity over longer periods and effectively smooths out the severe anomalies, outperforming the  UKF$^{\star}$ by 10.3\% and the IMM-UKF$^{\star}$ by 23.1\%. Crucially, the standard deviation of the MA-UKF error ($\pm 28.8$\,m) is nearly $8\times$ lower than that of the nominal baseline UKF ($\pm 229.9$\,m). The severe variance in the baselines correlates directly with the divergence events seen in the plot. By modulating the UT weights, the MA-UKF creates a flexible uncertainty manifold that instantaneously inflates the covariance yielding a highly resilient estimate.

\subsection{Analysis of Learned Adaptation Strategy}
\begin{figure}[htbp]
    \centering
    \includegraphics[width=1.0\linewidth]{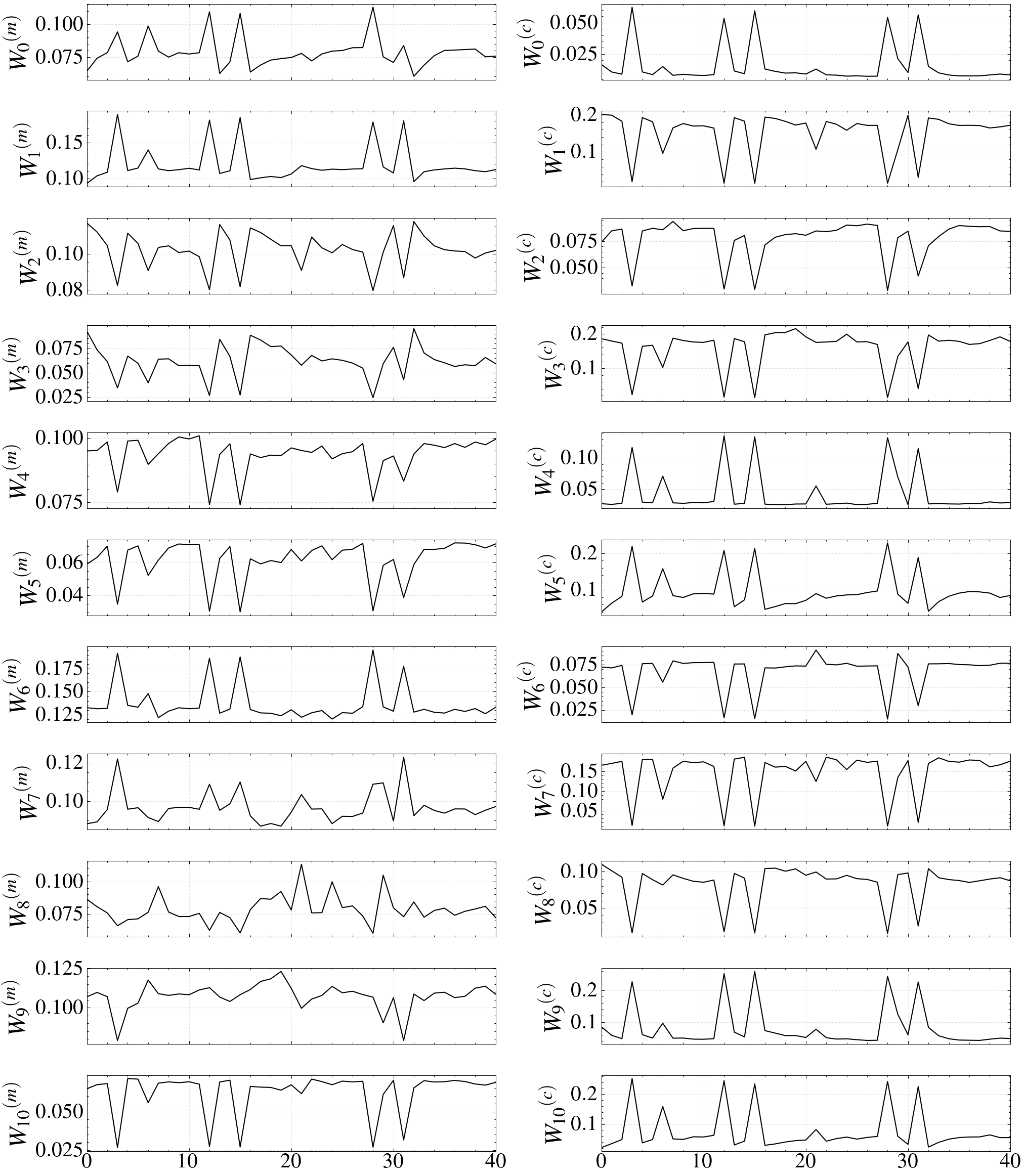}
    \caption{Temporal evolution of the learned sigma-point weights ($W_0$ to $W_{10}$) during the maneuver. The distinct impulsive spiking behavior across the entire geometry, contrasted with the continuous micro-modulation between events, indicates that the policy dynamically expands and contracts the local geometric spread to accommodate rapid directional changes and reject anomalies.}
    \label{fig:weights}
\end{figure}

To elucidate the decision-making logic of the meta-learner, we analyze the temporal evolution of the synthesized sigma-point weights. Fig.~\ref{fig:weights} displays the complete set of weight trajectories ($W_0$ through $W_{10}$) for both the mean ($W^{(m)}$) and covariance ($W^{(c)}$) distributions during a representative tracking sequence.

Contrary to the static parameterization of standard filtering (where UT scaling constants are strictly fixed), the MA-UKF exhibits a highly dynamic, context-dependent weighting strategy. Two distinct behavioral paradigms emerge from this analysis, reflecting the network's dual capacity for steady-state precision and transient robustness:

\begin{enumerate}
    \item \textbf{Continuous Micro-Modulation:}  Between major dynamic shifts, the entire set of mean and covariance weights represents a continuously shifting geometry, fluctuating dynamically around nominal baselines. This indicates that the filter actively and continuously compensates for localized linearization errors inherent to a fixed UT spread at every individual time step.  

    \item \textbf{Impulsive Covariance Resetting:}  Upon detecting unmodeled dynamic shifts or severe glint anomalies, the policy triggers a rapid reconfiguration of the sigma-point weights. As seen in Fig.~\ref{fig:weights}, the distribution maintains mostly flat baseline while triggering sharp, sparse spikes that strictly correlate with maneuver inflection points and outlier events. This coordinated, impulsive action injects structural uncertainty to accommodate unmodeled acceleration, increasing the Kalman gain's reactivity, while simultaneously leveraging the latent context embedding to gate out non-physical sensor noise.
\end{enumerate}

These observations visually confirm that the MA-UKF successfully internalizes a control policy. Rather than converging to static optimal parameters, it dynamically orchestrates the 11 pairs of UT weights as control inputs to continuously trade off trust between the process and measurement models, emulating an advanced adaptive filter within a fully continuous, differentiable architecture.

\section{Conclusion and Future Work}\label{sec:conclusions}

This paper introduced the Meta-Adaptive Unscented Kalman Filter (MA-UKF), a novel framework uniting rigorous Bayesian estimation with data-driven meta-learning. By casting the parameterization of the Unscented Transform as an end-to-end differentiable optimization problem, the MA-UKF learns to dynamically modulate its sigma-point geometry in real-time. This enables the filter to autonomously adapt to non-stationary dynamics and out-of-distribution regimes without relying on hand-crafted heuristics.

Utilizing a Recurrent Context Encoder to map historical innovation sequences into an actionable latent embedding, the proposed architecture effectively disentangles transient sensor anomalies from genuine target maneuvers. Empirical results on complex, out-of-distribution tracking scenarios under heavy-tailed glint noise demonstrate that the MA-UKF achieves robust generalization, reducing position ARMSE by 77.2\% and 10.3\% compared to the nominal and rigorously hyperparameter-optimized UKF baselines, respectively.

Future work will focus on validating the MA-UKF with real-world sensor data to assess Sim-to-Real transferability, and extending the differentiable formulation to Lie Groups to support robust 3D pose and orientation estimation in aerospace applications.

\bibliographystyle{IEEEtran}
\bibliography{main}

\end{document}